\documentclass[letterpaper, 10 pt, conference]{ieeeconf}  
\pdfoutput=1

\IEEEoverridecommandlockouts                              

\usepackage[backend=biber,style=ieee,citestyle=numeric-comp]{biblatex}
\usepackage{hyperref}
\usepackage{amsmath}
\usepackage{amssymb}

\usepackage{epsfig}
\usepackage{xcolor}
\usepackage{graphicx}
\usepackage{caption}
\usepackage{multirow}
\usepackage{booktabs}
\usepackage{cancel}
\title{\LARGE \bf
Reuse your features: unifying retrieval and feature-metric alignment
}

\author{Javier Morlana and J.M.M. Montiel
\thanks{This work was supported by the EU-H2020 grant 863146: ENDOMAPPER, the Spanish government grants PGC2018-096367-B-I00,  and by Arag\'on government grant DGA\_T45-17R.}
\thanks{The authors are with the Instituto de Investigaci\'on en Ingenier\'ia de Arag\'on (I3A), Universidad de Zaragoza, Mar\'ia de Luna 1, 50018 Zaragoza, Spain. E-mail: \{jmorlana, josemari\}@unizar.es.} }

\begin{document}

\maketitle

\begin{abstract}

We propose a compact pipeline to unify all the steps of Visual Localization: image retrieval, candidate re-ranking and initial pose estimation, and camera pose refinement. Our key assumption is that the deep features used for these individual tasks share common characteristics, so we should reuse them in all the procedures of the pipeline. Our DRAN (Deep Retrieval and image Alignment Network) is able to extract global descriptors for efficient image retrieval, use intermediate hierarchical features to re-rank the retrieval list and produce an initial pose guess, which is finally refined by means of a feature-metric optimization based on learned deep multi-scale dense features. DRAN is the first single network able to produce the features for the three steps of visual localization. DRAN achieves competitive performance in terms of robustness and accuracy under challenging conditions in public benchmarks, outperforming other unified approaches and consuming lower computational and memory cost than its counterparts using multiple networks. Code and models will be publicly available at \href{https://github.com/jmorlana/DRAN}{github.com/jmorlana/DRAN}.

\end{abstract}

\section{Introduction}

\label{section:intro}
\noindent Feature extraction is a relevant step in most computer vision tasks. Traditional approaches rely on image gradients to extract \emph{sparse features}, for example, edges or keypoints in a fixed hand-crafted manner. In the last decade, deep learning has taken the spot in feature extraction, with Convolutional Neural Networks (CNNs) as the most successful method. CNNs apply a set of convolutional filters to the image, obtaining a \emph{dense hierarchy of features}. Inherently, CNNs go deeper as the resolution decreases while obtaining higher semantics,  obtaining a pyramidal representation of the image.

We focus on visual localization for Visual SLAM (Simultaneous Localization And Mapping), i.e, we are assuming that a 3D map of the scene is available, which has been built using SfM (Structure-from-Motion) or Visual SLAM. The map is composed of 3D points, $\mathbf{P}_i$, and keyframes. Keyframes, or reference frames, are a selected set of images from which the map geometry is computed by Bundle Adjustment. Per each keyframe, we have available its image $\mathbf{I}_j$ and its camera pose $\left\{\mathbf{R}_j,\mathbf{t}_j\right\}$. Given a query image $\mathbf{I}_q$ and the map, the camera location procedure efficiently retrieves the closest map keyframes $\mathbf{I}_k$ and estimates the 6-DoF pose of the query image with respect to the map, $\left\{\mathbf{R}_q,\mathbf{t}_q\right\}$. 


First we apply a \emph{keyframe retrieval step}, in which the typical deep learning retrieval algorithm usually takes three sub-steps: i) the query image is forwarded through the network encoder, obtaining dense hierarchical features, ii) the deepest feature map is pooled \cite{radenovic2018fine, arandjelovic2016netvlad} obtaining a compact global descriptor, and iii) the descriptor is compared against the other global descriptors of the keyframes in the database, ranking the keyframes by descriptor similarity. Images are translated into compact vectors based on the deepest features, which typically encodes the high-level features of the image. The comparison of global descriptors gives an initial list of candidate keyframes ranked by similarity. 

For the \emph{camera feature-metric pose estimation step}, we
apply deep image alignment techniques \cite{sarlin2021back, von2020gn, von2020lm, lv2019taking} minimizing a feature-metric error of the dense hierarchical features that encode the image at different resolutions, estimating the relative pose between the query and a keyframe. These methods are robust against illumination and point of view changes, and achieve high precision as they can get subpixel accuracy, but they need a good initial pose guess to converge to the correct minimum.

Most of the candidate keyframes in the list depict the same place as the query and might provide a coarse pose initialization to the feature-metric optimization, but their sorting criteria does not take into account the conditions that most favour the pose optimization, i.e. point of view similarity. For this reason, the intermediate \emph{re-ranking step} re-sort the keyframes in the initial list to prioritize the overlap and point of view similarity with the query. This step exploits the dense hierarchy of features to produce matches between the query and the keyframes, yielding 2D-3D matches between the query and the map. The matches are refined by a PnP+RANSAC, obtaining an initial pose guess which is closer to the true one, and the feature-metric optimization can successfully converge.

Other works in literature deal with Visual Localization employing different networks for each of the steps. In contrast to them, we use a unique dense hierarchy of features, what is efficient and elegant, providing competitive results of accuracy. Benefits of unified approaches are threefold. Firstly, unified learnt approaches will lead the way to deep direct SLAM algorithms able to use their own features for relocalization and loop closure. Direct SLAM methods use image intensities to perform tracking but they need additional features for relocalization and loop closure. Secondly, unified approaches could benefit from the image retrieval training data, which is typically easier to obtain as it is labelled only at the image level, using this same data for the training of their local descriptors. And lastly, it is obviously more efficient as the images are processed by a single network.

\begin{figure*}[ht]
  \centering
   \includegraphics[width=0.78\linewidth]{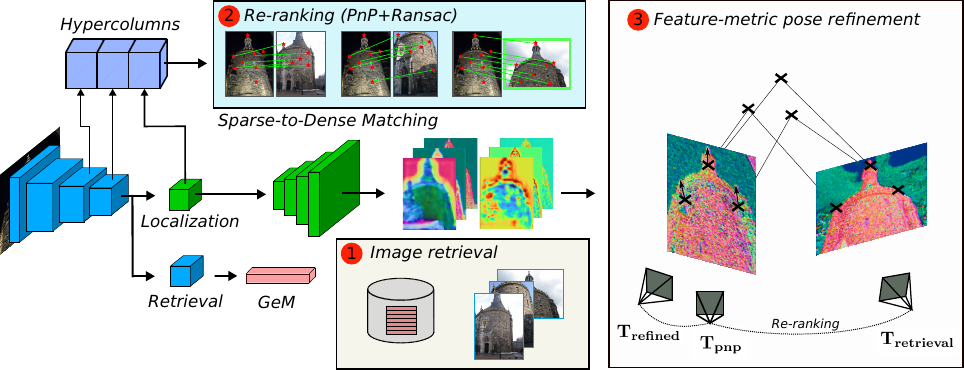}
   \vspace{-0.1cm}
    \caption{\textbf{DRAN pipeline.} DRAN provides the features for the three steps of Visual Localization with a single network. The architecture follows a U-Net style with two heads: localization and retrieval. The green blocks are trained specifically for feature-metric alignment, while the blue ones  are pretrained in a generic retrieval dataset and frozen during training.}
    
   \label{fig:dran}
   \vspace{-0.15cm}
\end{figure*}

Our key insight is that it is beneficial to use a unique hierarchy of learned features for all the steps of the camera visual localization. Following these ideas, we propose DRAN (Deep Retrieval and image  Alignment  Network, Figure \ref{fig:dran}), a unified architecture that combines the knowledge from retrieval and camera pose stages. Our contributions are:

\begin{itemize}
    \item DRAN, a multi-task single network providing, in a single forward pass, both an image global descriptor for retrieval, hypercolumns for re-ranking and initial pose estimation, and a multi-level dense feature hierarchy for feature-metric camera pose refinement. 

    
    \item An evaluation under challenging conditions, showing that our system achieves better performance than other unified systems, and it is competitive against feature matching pipelines that combine different networks.
\end{itemize}

\vspace{-0.1cm}

\section{RELATED WORK}

\label{sec:related}
We review the relevant work in the related areas of this paper: direct and indirect SLAM, image retrieval, deep camera pose estimation and unified methods.

\textbf{Direct and Indirect SLAM} is the widely used classification for traditional SLAM techniques. Indirect SLAM \cite{mur2015orb,campos2020orb, davison2007monoslam, klein2007parallel},  processes the image to detect, describe and match a sparse set of keypoints that are robust to illumination and viewpoint changes, these matches are fed in a geometric Bundle Adjustment (BA) to recover the scene geometry. In contrast, direct SLAM operates with raw image intensities, relying on brightness consistency \cite{engel2017direct, engel2014lsd, newcombe2011dtam} to also feed, in this case, a photometric BA. Direct alignment allows sub-pixel accuracy but is vulnerable to illumination changes and suffers from small convergence basin. We propose the use of learned features to overcome the challenges of direct methods when applied to camera pose estimation. One of the first works attempting this is \cite{gladkova2021tight}, which proposes the integration of learned features for relocalization into DSO \cite{engel2017direct}. Differently, a SLAM system with our unified approach as a feature extractor would use its own features for every task, without relying on classical photometric alignment.

\textbf{Image retrieval (IR)} stands for the task of efficiently retrieving an image among a database of visited places, in our case the keyframes. To perform a quick search, a compact image representation is needed. Traditionally, the image embedding was obtained by the aggregation of hand-crafted local descriptors such as SIFT \cite{lowe2004distinctive}, ORB \cite{rublee2011orb} in a Bag-of-Words \cite{Nister2006,GalvezTRO12} representation. Nowadays, the field is dominated by CNN representations that aggregate feature maps into a global descriptor \cite{arandjelovic2016netvlad, radenovic2018fine, revaud2019learning, radenovic2016cnn, teichmann2019detect}. IR training only requires image-level labels, i.e. if two images depict the same place (positives) or not (negatives). This is much easier to obtain than precise pixel-level labels, which are typically needed for deep local features. The global descriptor allows two images to be compared efficiently with a single distance computation. Our work adopts the state-of-the-art pooling method Generalized-Mean (GeM) for the IR step\cite{radenovic2018fine}.

\textbf{Deep camera pose estimation} mainly encompasses three approaches: pose regression, image matching and deep direct alignment. Pose regression \cite{kendall2017geometric, naseer2017deep} learns to directly map the input to pose parameters without any 3D constraint. They require a lot of training data and do not generalize well to novel domains. Differently, image matching extracts local features \cite{detone2018superpoint, dusmanu2019d2, sarlin2019coarse} and performs data association by descriptor distance or learned matchers \cite{sarlin2020superglue, sun2021loftr}. The pose can be obtained with a PnP \cite{lepetit2009epnp} algorithm if those local features are present in a 3D reference model. Deep direct alignment \cite{von2020gn, sarlin2021back, lv2019taking, xu2020deep, von2020lm} takes an approach similar to direct SLAM, trying to minimize a photometric error based on the deep features of a CNN, i.e. a feature-metric optimization. In this work, we build on top of the recent framework PixLoc \cite{sarlin2021back}, which extracts multi-scale dense descriptors that are aligned iteratively in a pyramidal approach. The feature-metric error is evaluated on the 2D projections of the corresponding 3D model built by SfM. S2DNet \cite{germain2020s2dnet} is another approach that also takes advantage of the map projections to match sparse deep descriptors from the reference frame against the dense descriptors of the query. In contrast to us, S2DNet is trained to solve only the matching step between two images.

\textbf{Unified approaches} try to join the local and global descriptor extraction into a single system. DELG \cite{cao2020unifying} obtains a  GeM global descriptor \cite{radenovic2018fine} and applies an attention mechanism \cite{noh2017large} to obtain local descriptors for re-ranking the initial candidates. HF-Net \cite{sarlin2019coarse} proposed a distillation framework in a teacher-student approach to learn from NetVLAD \cite{arandjelovic2016netvlad} and SuperPoint \cite{detone2018superpoint}, being able to perform retrieval and camera localization. UR2KiD \cite{yang2020ur2kid} also performs retrieval and matching, with the benefit of being trained only with image labels. S2DHM \cite{germain2019sparse} uses an encoder pretrained for retrieval to perform re-ranking and extract an initial pose under challenging conditions. We use the ideas of S2DHM but applying them to networks trained in generic retrieval datasets (SfM120k \cite{radenovic2018fine}, MSLS \cite{warburg2020mapillary}). Besides, differently from them, our work unifies the tasks of image retrieval, local matching and feature-metric image alignment, introducing, to the best of our knowledge, the first system to combine them in the same network.

\vspace{-0.4cm}


\section{Features for Retrieval and Alignment}

\vspace{-0.3cm}



We propose the DRAN (Deep Retrieval and image Alignment Network) architecture, whose main steps are described in Fig. \ref{fig:dran}. We argue that most of the meaningful features are already extracted by IR networks, so we do not need to train another full pipeline end-to-end to detect features for the camera pose estimation. We adopt a shared encoder architecture with two heads: retrieval and localization. The encoder is a VGG16 net pretrained for IR in the SfM120k/MSLS dataset, which is splitted after the \texttt{conv4} stage. We use retrieval datasets to pretrain our encoder, as this data is easier to obtain than accurate 3D models and allows to acquire invariance against challenging conditions.

The retrieval head incorporates the last stage of the truncated encoder (\texttt{conv5}) and a GeM pooling layer that aggregates the last feature map into a compact vector. As we do not want to affect the retrieval performance, both the encoder and the retrieval head remain frozen during training, retaining their original weights fine-tuned for retrieval.

The localization head has identical structure as the retrieval head (\texttt{conv5} of VGG16), with the difference that it is optimized during training. It connects to a decoder network with skip connections in a U-Net \cite{ronneberger2015u} style. Following \cite{sarlin2021back}, we optimize all the weights involved in order to learn features for accurate camera localization (\ref{subsec:align}). The output of the decoder is the hierarchy of features, along with its uncertainty. The representation of an image for each scale level $l$ is a feature map $\mathbf{F}^{l} = \mathbb{R}^{W_{l} \times H_{l} \times D_{l}}$ and its per pixel uncertainty $\mathbf{U}^l=\mathbb{R}^{W_{l} \times H_{l}}$. The Levenberg-Marquardt optimization will minimize the feature-metric error in these features, as explained in \ref{subsec:align}.

To provide a good initial guess for the feature-metric camera pose optimization, we extract hypercolumns \cite{germain2019sparse} using the features computed by the shared encoder and the localization head. We filter the retrieval candidates and obtain an initial pose by means of a PnP+RANSAC. This pose is better than the coarse pose obtained by IR, boosting the localization performance of the subsequent optimization.

\vspace{-0.4cm}

\subsection{Compact Global Image Descriptor}
\label{subsec:global}
\vspace{-0.3cm}

We adopt the Generalized-Mean (GeM) \cite{radenovic2018fine} pooling layer as the method to aggregate the activations from the last layer of the retrieval head. The GeM operation for each channel of the $C\times H\times W$ activation maps is described in Eq. \ref{eq:gem}. $\mathcal{X}_k$ is the set of $HW$ activations of each channel and $p_k$ is the learnt parameter that controls the pooling operation.

\vspace{0.3cm}
\begin{equation}
  \mathbf{f} =\left[f_1 \ldots f_k \ldots f_K\right]^\top \in \mathbb{R}^{512},\;\;f_k = \left ( \frac{1}{\left | \mathcal{X}_k \right |} \sum_{x \in \mathcal{X}_k} x^{p_k}  \right )^{\frac{1}{p_k}} 
  \label{eq:gem}
\end{equation}

 Both the encoder and $p_k$ are trained on the SfM120k \cite{radenovic2018fine} dataset, which depicts popular landmarks, or in MSLS \cite{warburg2020mapillary}, composed of urban and suburban scenes. It is trained using a siamese architecture and the contrastive loss \cite{radenovic2018fine, leyvavallina2021gcl}. The contrastive loss takes as input a tuple containing 1 query, 1 positive example and 5 negative examples. Here the objective is to obtain similar descriptors for images depicting the same place, where the distance is computed by the dot product of two L2-normalized global descriptors. For testing, multiscale and whitening is applied as in \cite{radenovic2018fine}.

\vspace{-0.2cm}

\subsection{Re-ranking and initial pose estimation}
\label{subsec:rerank}

\begin{figure}[t]
    \centering
    \includegraphics[width=0.92\columnwidth]{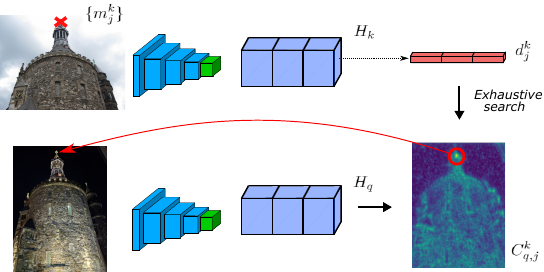}
    \vspace{-0.15cm}
    \caption{Exhaustive matching for re-ranking and initial pose. Only points in the 3D map are matched, obtaining a sparse set of descriptors from the dense hypercolumn $H_k$. Each descriptor $d_{j}^{k}$ is searched densely in the query hypercolumn $H_q$, identifying the match as the global maxima of $C_{q,j}^{k}$.}
    \label{fig:hypercolumns}
    \vspace{-0.7cm}
\end{figure}

The IR module focuses on finding keyframes depicting the same place than the query, but in some cases, the first candidate is not the best initialization for feature-metric alignment. Images can suffer from big changes in scale or in point of view, or little overlap, resulting in a bad pose initialization that hinders convergence to the optimal pose.

For this reason, we employed a re-ranking method that minimize this issue by selecting the best candidate to initialize with, based on the number of inliers. We brought the method S2DHM (Sparse-to-Dense Hypercolumn Matching)  proposed by Germain et al. \cite{germain2019sparse} to our features. The goal is to obtain local matches between the query and the keyframes with the encoder features, in order to filter the image retrieval candidates and estimate an initial 6DoF pose guess.

Given a query $I_{q}$ and a keyframe $I_{k}$, we extract features from layers  \texttt{conv\_3\_3},  \texttt{conv\_4\_1},  \texttt{conv\_4\_3},  \texttt{head\_1} and  \texttt{head\_3}, using our trained localization head instead of the pretrained for retrieval. These features are upsampled using bilinear interpolation to match with the resolution of the earliest layer (\texttt{conv\_3\_3}, resolution is 1/4 of the input image) and concatenated. The result is the so-called \textit{hypercolumns} $H_q$ and $H_k$, a dense descriptor that encodes information from different levels of the network.

As we already have a 3D map, we will only try to match the 3D points $P_k$ that were already detected in 2D locations $\{m_{j}^{k}\}_{j=1\ldots P_{k}}$ in the keyframe. We interpolate the hypercolumn $H_k$ at each location $\{m_{j}^{k}\}$, obtaining a sparse set of descriptors $\{d_{j}^{k}\}_{j=1\ldots P_{k}}$. To find the matches between the dense query hypercolumn and the sparse keyframe descriptors, a dot product is computed between every sparse descriptor $d_{j}^{k}$ and the dense hypercolumn, obtaining a cross-correlation map $C_{q,j}^{k} = H_q \cdot d_{j}^{k}$. The tentative matches corresponds to the global maximum of the cross-correlation map for each of the points $p_{k}$ considered. To avoid outliers due to repetitive patterns or occlusions, a ratio test is conducted.

The resulting $P_{k}$ 2D-3D matches are fed into a PnP+RANSAC scheme, which outputs the final inliers and the initial pose estimation. The keyframe with the most inliers is chosen, along with the PnP pose estimated for the query. This pose is the seed given to the final optimization.

\vspace{-0.2cm}
\subsection{Deep direct feature-metric image alignment}
\label{subsec:align}

\begin{table*}[t]
\centering
\resizebox{0.97\textwidth}{!}{%
    \centering
    \begin{tabular}{clccccccc}
        \toprule
        & \multirow{2}{*}{Method} & \multicolumn{2}{c}{Aachen Day-Night} & \multicolumn{2}{c}{RobotCar Seasons} & \multicolumn{3}{c}{Extended CMU Seasons} \\ \cmidrule(lr){3-4} \cmidrule(lr){5-6} \cmidrule(lr){7-9}
                            &    & Day               & Night             & Day               & Night                 & Urban     & Suburban      & Park        \\

        \midrule
        \parbox[t]{2mm}{\multirow{4}{*}{\rotatebox[origin=c]{90}{\textbf{\textcolor{red}{FM}}}}} & Pixloc                  & 64.3 / 69.3 / 77.4    & 51.0 / 55.1 / 67.3   & 52.7 / 77.5 / 93.9  &  12.0 / 20.7 / 45.4   & 88.3 / 90.4 / 93.7   & 79.6 / 81.1 / 85.2    &  61.0 / 62.5 / 69.4\\
        & S2DNet                  & 84.3 / 90.9 / 95.9    & 46.9 / 69.4 / 86.7   &  53.9 / 80.6 / 95.8  &  14.5 / 40.2 / 69.7   & -   & -    &  -\\
        & D2-Net                  & 84.3 / 91.9 / 96.2    & 75.5 / 87.8 / 95.9   & 54.5 / 80.0 / 95.3  &  20.4 / 40.1 / 55.0   & 94.0 / 97.7 / 99.1   & \textbf{\textcolor{red}{93.0}} / \textbf{\textcolor{red}{95.7}} / \textbf{\textcolor{red}{98.3}}    &  \textbf{\textcolor{red}{89.2}} / \textbf{\textcolor{red}{93.2}} / \textbf{\textcolor{red}{95.0}}\\
        & hloc                    & \textbf{\textcolor{red}{89.6}} / \textbf{\textcolor{red}{95.4}} / \textbf{\textcolor{red}{98.8}}    & \textbf{\textcolor{red}{86.7}} / \textbf{\textcolor{red}{93.9}} / \textbf{\textcolor{red}{100}}   & \textbf{\textcolor{red}{56.9}} / \textbf{\textcolor{red}{81.7}} / \textbf{\textcolor{red}{98.1}} & \textbf{\textcolor{red}{33.3}} / \textbf{\textcolor{red}{65.9}} / \textbf{\textcolor{red}{88.8}} & \textbf{\textcolor{red}{95.5}} / \textbf{\textcolor{red}{98.6}} / \textbf{\textcolor{red}{99.3}} & 90.9 / 94.2 / 97.1 & 85.7 / 89.0 / 91.6\\
        
        \midrule
        \parbox[t]{2mm}{\multirow{4}{*}{\rotatebox[origin=c]{90}{\textbf{\textcolor{blue}{Unified}}}}} & S2DHM                   & 56.3 / 72.9 / 90.9    & 30.6 / 56.1 / 78.6   & 45.7 / 78.0 / \textbf{\textcolor{blue}{95.1}} &  \textbf{\textcolor{blue}{22.3}} / \textbf{\textcolor{blue}{61.8}} / \textbf{\textcolor{blue}{94.5}}  & 65.7 / 82.7 / 91.0 & 66.5 / 82.6 / \textbf{\textcolor{blue}{92.9}} & 54.3 / \textbf{\textcolor{blue}{71.6}} / \textbf{\textcolor{blue}{84.1}}\\
        & HF-Net                  & \textbf{\textcolor{blue}{79.9}} / 88.0 / 93.4    & 40.8 / 56.1 / 74.5   & 53.0 / 79.3 / 95.0  &  5.9 / 17.1 / 29.4  & \textbf{\textcolor{blue}{89.5}} / \textbf{\textcolor{blue}{94.2}} / \textbf{\textcolor{blue}{97.9}} & 76.5 / 82.7 / 92.7 & 57.4 / 64.4 / 80.4\\
        & UR2KiD                  & \textbf{\textcolor{blue}{79.9}} / \textbf{\textcolor{blue}{88.6}} / \textbf{\textcolor{blue}{93.6}}    & 45.9 / 64.3 / 83.7  & -   & -    &  - & -   & -\\
        & \textbf{DRAN} (ours)    & 76.9 / 86.2 / 90.8    & \textbf{\textcolor{blue}{65.3}} / \textbf{\textcolor{blue}{78.6}} / \textbf{\textcolor{blue}{85.7}}  & \textbf{\textcolor{blue}{55.9}} / \textbf{\textcolor{blue}{80.7}} / \textbf{\textcolor{blue}{95.1}} &	19.8 / 36.7 / 54.8  &  88.7 / 91.4 / 93.8 &  \textbf{\textcolor{blue}{85.4}} / \textbf{\textcolor{blue}{87.5}} / 90.0  &  \textbf{\textcolor{blue}{67.3}} / 69.7 / 72.0\\
        
        
        \bottomrule
    \end{tabular}}
    \vspace{-0.1cm}
    \caption{Results for Large-scale localization in Aachen Day-Night, RobotCar Seasons and Extended CMU Seasons. We highlight in red the best result for the \textbf{\textcolor{red}{Feature Matching}} alternatives, while in blue the best result for the \textbf{\textcolor{blue}{Unified}} algorithms.}
    \label{table:aachen}
    \vspace{-0.2cm}
\end{table*}

We employ the method proposed in \cite{sarlin2021back} to align our hierarchy of deep features. As other recent works \cite{von2020gn, lv2019taking, von2020lm, xu2020deep}, they treat deep image alignment as a direct feature-metric optimization problem, in a similar way as DSO \cite{engel2017direct} minimizes the photometric error. We describe it briefly, for further details please refer to \cite{sarlin2021back}.

As PixLoc proposes, we extract $L$=3 feature maps from the U-Net decoder, with strides 1, 4 and 16. The shallow levels encode low texture cues while deeper levels capture high level features and semantic content. This pyramidal representation is similar to traditional photometric alignment. Learning this representation instead of relying on the raw photometric values allows to overcome with the known limitations of direct image alignment: illumination changes and small convergence basin. For a feature level $l$, the feature-metric residual is defined as weighted addition of the feature-metric error between the query and the retrieved images for the 3D map points detected in the query image:
\begin{eqnarray}
    \vspace{-0.4cm}
    E_{l}(\mathbf{R}_q, \mathbf{t}_q) &=& \sum_{i,k} w_{k}^{i}\rho(\left \| \mathbf{r}_{k}^{i}\right \|_{2}^2)
    \label{eq:error}\\
    \mathbf{r}^{i}_{k} &=& \mathbf{F}^{l}_{q} \left [  \mathbf{p}_{q}^i \right ] - \mathbf{F}^{l}_{k} \left [  \mathbf{p}_{k}^i\right ] \in \mathbb{R}^{D}
    \label{eq:residual} \\
w_k^i &=& \frac{1}{1+\mathbf{U}^l_q\left[ \mathbf{p}_{q}^i\right]} \frac{1}{1+\mathbf{U}^l_k\left[ \mathbf{p}_{k}^i\right]} \in [0,1]
\label{eq:per-pixel-uncertainty}
\end{eqnarray}

Where $\mathbf{F}^{l}_{q}$ and $\mathbf{F}^{l}_{k}$ are the feature maps for the query and the keyframe at a certain level $l$, $\left [ \mathbf{p}_{q}^i \right ]$ and $\left[  \mathbf{p}_{k}^i \right ]$ are the projections of the point $\mathbf{P}_{i}$ with subpixel accuracy, and $\mathbf{U}^l_q$ and $\mathbf{U}^l_k$ are the predicted uncertainty maps. $w_k^i$ learns to determine if the location of a 3D point is good for localization or not. If the point projection has low uncertainty in both the query and the reference image, $w_k^i$ will tend to 1. Otherwise, $w_k^i$ will tend to 0, weighting down the residual in the optimization.
For $\mathit{N}$ points, (\ref{eq:error}) defines the goal function to be minimized with the Levenberg-Marquadt (LM) algorithm. The network learns to find good features to localize and whether a point is reliable or not. 

The initial guess for the iterative optimization is the one selected by the Sparse-to-Dense matching. The algorithm starts optimizing the coarsest level (stride 16), the feature maps with lower resolution but higher depth, and successively optimizes the finer levels, going from a coarse estimation to a finer one. As the finest layer has the same resolution as the input image, feature-metric optimization can achieve subpixel accuracy, just like classic photometric alignment. The LM optimization for the camera pose comes down to solve the linear system $- \left(\mathbf{H} + {\boldsymbol\lambda}\, \mbox{diag}\left(\mathbf{H}\right)\right){\boldsymbol \delta} = \mathbf{J}^\top\mathbf{W}\mathbf{r}$, where $\mathbf{H}=\mathbf{J}^\top\mathbf{W}\mathbf{J}$. $\mathbf{J}$ is the Jacobian, $\mathbf{W}$ is weighting matrix depending on (\ref{eq:per-pixel-uncertainty}) and $\boldsymbol \delta \in \mathbf{SE}(3)$ is the pose update parameterized by its Lie algebra. The damping parameter $\boldsymbol \lambda$ is formulated as a fixed and learned $6\times6$ diagonal matrix coding the damping independently in each of the 6 DoF of the camera pose and for each of the levels.




\textbf{Loss function} The only supervision for learning is the ground truth pose for the query images,
$\left\{ \bar{\mathbf{R}}_q,\bar{\mathbf{t}}_q \right\}$. The loss function (\ref{eq:reprojection}) penalizes the Huber cost of distance in pixels the between the reprojection of the map points in the ground truth camera pose, and the reprojection of the same points in the feature-metric optimized camera pose,$\left\{ \mathbf{R}_{l,q}, \mathbf{t}_{l,q} \right\}$, averaging among the different scales. 
\begin{equation}
    \mathcal{L} = \frac{1}{L} \sum_{l} \sum_{i}  \left \| \Pi(\mathbf{R}_{l,q}\mathbf{P}_{i}+\mathbf{t}_{l,q}) - \Pi(\bar{\mathbf{R}}_{q}\mathbf{P}_{i}+\bar{\mathbf{t}}_{q}) \right \|_{\gamma}
    \label{eq:reprojection}
    \vspace{-0.1cm}
\end{equation}
Where $\Pi$ represents the reprojection transformation and $\gamma$ is the Huber cost. This formulation is not affected by the geometric scale of the scene, as it only works with the reprojection of the points and not directly with camera poses.


\section{Experiments}

In this section we evaluate the advantages of our \textit{retrieval deep alignment}, comparing it against other learned approaches. In section \ref{subsec:datasets}, we explain the datasets used for training and evaluation, and the baselines used to compare. We show the accuracy of the camera pose estimated by our system in large-scale visual localization in section \ref{subsec:large_scale}, performing better than the other unified approaches and competitively against feature matching approaches. Finally, in section \ref{subsec:ablation}, we perform an ablation study, showing the benefits of each of the elements of the pipeline, and the run time and memory efficiency of the DRAN.

\vspace{-0.1cm} 





\subsection{Datasets and baselines}
\label{subsec:datasets}

\textbf{Training} As other works \cite{sarlin2021back, germain2019sparse}, we train two models in different datasets to learn specific domains. For the first one, our retrieval encoder is pretrained on SfM120k \cite{radenovic2018fine}, which depicts common landmarks around the world and clustered with COLMAP \cite{schonberger2016structure}. Training uses the contrastive loss and hard-negative mining. We experimented with the original version given by the authors, which initializes their training with weights from custom ImageNet pretraining on Caffe \cite{jia2014caffe}, but we found better convergence when initializing with PyTorch \cite{paszke2019pytorch} ImageNet pretraining \cite{deng2009imagenet}. We use MegaDepth \cite{li2018megadepth} dataset for training the feature-metric alignment procedure. It contains about 1 million images depicting popular landmarks, grouped into 196 scenes and reconstructed by COLMAP \cite{schonberger2016structure}. MegaDepth provides depth maps and pose information for every camera. We use the same split as D2-Net \cite{dusmanu2019d2} for training and validation. The training is performed for 20k iterations with the Adam optimizer, using a constant learning rate of $5 \times 10^{-6}$ and a batch size of 6. This first model will be evaluated in Aachen Day-Night.

For our second model, our retrieval encoder is pretrained on MSLS \cite{warburg2020mapillary}, a large dataset for urban and suburban place recognition from images sequences. We used the model provided by \cite{leyvavallina2021gcl}, which was trained in MSLS using their Generalized Contrastive Loss. As PixLoc \cite{sarlin2021back}, we use the training set of the Extended CMU dataset for the feature-metric training. The training is performed for 40k iterations with the Adam optimizer, using a constant learning rate of $1 \times 10^{-5}$ and a batch size of 3. This model will be evaluated in RobotCar and the Extended CMU seasons datasets.


\begin{figure*}[t]
    \centering
    \includegraphics[width=0.93\linewidth]{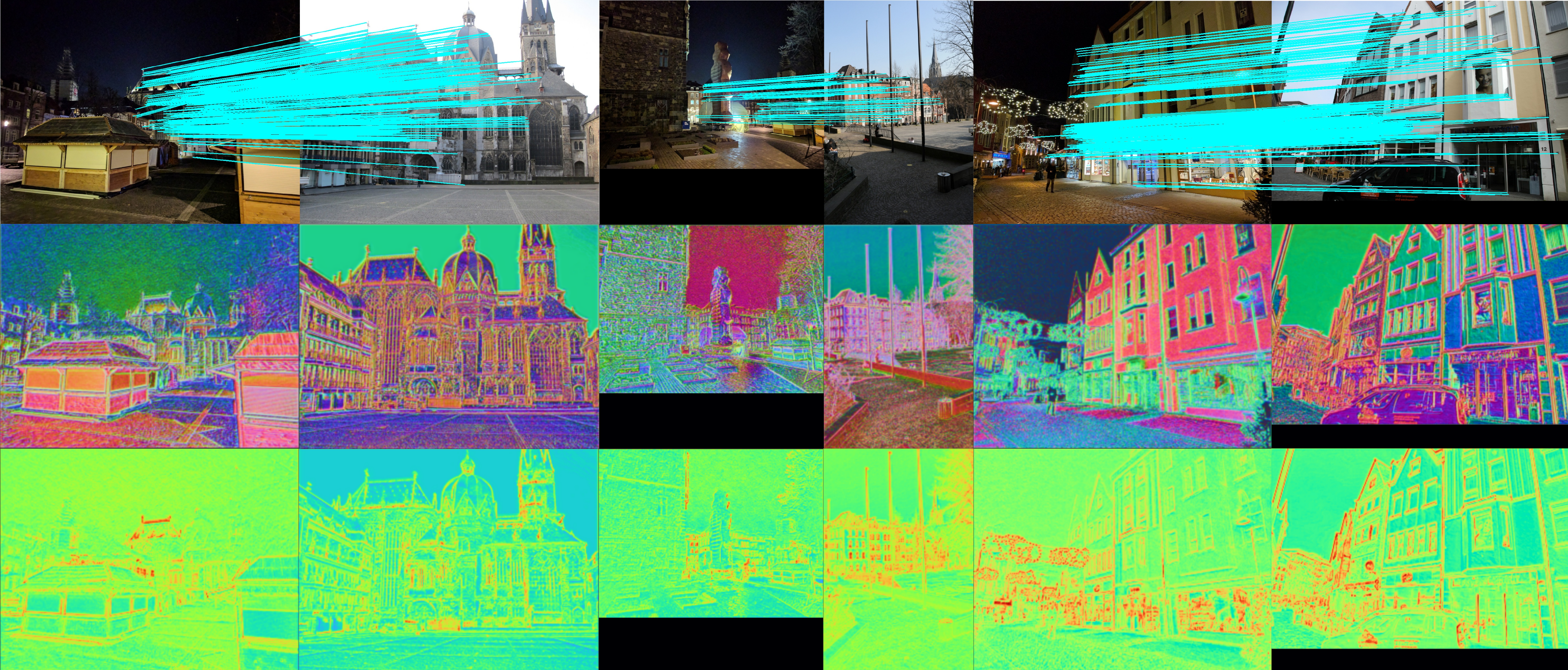}
    \vspace{-0.1cm}
    \caption{Examples from Aachen Day-Night. Each two columns depicts a query-reference pair. The first row shows the Sparse-to-Dense matching of our approach, showing impressive results in night conditions under severe viewpoint change or occlusions. The second row shows the features $\mathbf{F}$ from the finest resolution learned in the decoder. The last row shows the also learned uncertainty maps $\mathbf{U}$, where red means that points are more reliable, and blue that points are ignored.  }
    \label{fig:matching}
\end{figure*}

\textbf{Baselines} We compare against two groups of methods: Feature Matching (FM) and Unified methods. 

For Feature Matching, we consider methods for matching between two images, where the retrieval is given by an external IR network, typically NetVLAD. In FM, we find the state-of-the-art methods for local matching as D2-Net \cite{dusmanu2019d2} or the toolbox hloc \cite{sarlin2019coarse, sarlin2020superglue}, which uses SuperPoint \cite{detone2018superpoint} and SuperGlue \cite{sarlin2020superglue}. We also compare against PixLoc \cite{sarlin2021back}, which performs feature-metric optimization, and S2DNet \cite{germain2020s2dnet}, a learned matching system for sparse-to-dense matching.

In the Unified methods, we consider methods able to perform retrieval and camera pose estimation with a single network. Here we found S2DHM \cite{germain2019sparse}, which performs sparse-to-dense matching with an encoder trained for retrieval in a subset of RobotCar Seasons or Extended CMU. HF-Net \cite{sarlin2019coarse} proposes a pipeline that distilles knowledge from NetVLAD \cite{arandjelovic2016netvlad} and SuperPoint \cite{detone2018superpoint} in a compact network. UR2KiD\cite{yang2020ur2kid} is able to perform retrieval and local matching while only being trained with image-level labels (no keypoints needed).



\vspace{-0.1cm}

\subsection{Large-scale localization}
\label{subsec:large_scale}



\textbf{Aachen Day-Night} objective is to evaluate visual localization under challenging conditions. It is composed of 4,328 reference images depicting the city of Aachen, taken during daytime with hand-held devices. A 3D model is reconstructed with these images, and 922 queries (824 daytime, 98 nighttime) along with its 6DoF are provided. The benchmark protocol from \cite{sattler2018benchmarking} reports the percentage of queries localized for different thresholds for the camera position and orientation error. Localization recall is provided for three threshold levels: (25\,cm, 2º), (50\,cm, 5º) and (5\,m, 10º), which can be seen in Table\,\ref{table:aachen}. For obtaining the retrieval candidates with DRAN, we use multiscale and learned whitening, for scales $1, \frac{1}{\sqrt{2}}$ and $\frac{1}{2}$, as proposed by \cite{radenovic2018fine}.

Among the unified approaches, DRAN performs the best by a great margin in the night-queries, while performing comparably as UR2KiD in the day ones. Feature matching pipelines as hloc, that uses NetVLAD, SuperPoint and SuperGlue, perform better than ours, with the drawback that they use several networks to perform localization, while we only need one. DRAN obtains the camera pose estimation using $\textit{N} = 5$ retrieval candidates. hloc, for example, reports results using 50 candidates.

Aachen Day-Night reference poses are really sparse, providing a extremely coarse initialization that makes difficult to pure feature-metric methods to converge. Here is where the sparse-to-dense method shines, finding matches in difficult situations (Figure \ref{fig:matching}) and providing a much more precise initial pose to the feature-metric optimization. In contrast to PixLoc, which only performs the optimization with a given retrieval, we are able to perform our own retrieval and estimate more accurate and robust camera poses.

\textbf{RobotCar Seasons and Extended CMU Seasons} show two driving scenarios. RobotCar is comprised of several sequences under different weather conditions  that are classified under day and night groups. CMU is comprised by three kind of scenarios: urban, suburban and park, which are depicted under seasonal changes. We use multiscale and PCA whitening  to obtain the retrieval candidates.

For RobotCar, our method outperforms all the other Unified approaches in the Day condition and obtains similar results as hloc, the best Feature Matching method. In the Night condition, we perform similarly as S2DHM in the finest threshold, but they outperform us in the others. The high night-time performance of S2DHM could be explained because it is the only method that has been trained in a subset of RobotCar Seasons, including night images.

For Extended CMU, the results are heterogeneous. Complex Feature Matching pipelines as hloc and D2-Net perform better than Unified approaches, with the drawback of needing an external retrieval. We outperform the other Unified frameworks at the finest level in Suburban and Park scenes, and perform comparably to HF-Net in the Urban set. This higher performance in the finest threshold arguably comes from the feature-metric optimization, being DRAN the only Unified approach that applies this step. We outperform PixLoc, the other pure feature-metric approach, in all of the scenes.

\vspace{-0.1cm}



\subsection{Ablation study and Runtime}
\label{subsec:ablation}

\begin{table}[t]
\resizebox{0.90\columnwidth}{!}{%
    \centering
    \begin{tabular}{lcc}
        \toprule
        \multirow{2}{*}{Method} & \multicolumn{2}{c}{Aachen Day-Night} \\ \cmidrule(lr){2-3}
                                & Day               & Night            \\
        
        \midrule
        DRAN (R+A)          & 58.6 / 64.6 / 74.2   & 42.9 / 44.9 / 51.0   \\
        DRAN (R+P)          & 73.1 / 84.0 / 90.8   & 58.2 / 70.4 / 85.7   \\
        DRAN (R+P+A)        & 76.9 / 86.2 / 90.8   & 65.3 / 78.6 / 85.7   \\
        DRAN - light        & 76.1 / 85.6 / 90.3   & 63.3 / 74.5 / 83.7   \\
        \midrule
        GeM+S2DHM+PixLoc    & 78.0 / 86.7 / 91.9   & 66.3 / 75.5 / 89.8     \\
        
        
        \bottomrule
    \end{tabular}}
    \vspace{-0.1cm}
    \caption{Ablation study in Aachen Day-Night.}
    \label{table:ablation}
    \vspace{-0.2cm}
\end{table}


We perform an ablation study in the Aachen Day-Night dataset of the different modules of our algorithm in Table\,\ref{table:ablation}. (R+A) uses our top scored keyframe of our retrieval network (R) as the intial guess for deep image alignment (A), gives similar results on the day to Pixloc, the most comparable method, but lags in the night queries because the system can not overcome bad retrieval initializations.

    

Performing retrieval and only PnP with hypercolumns (R+P) allows to produce an accurate pose by itself giving a huge boost to the performance. Using all the above modules (R+P+A),  where the feature-metric optimzation uses R+P as initial guess conforms a unified pipeline that beats the state-of-the-art of multitasks methods in Aachen night queries, while being competitive on the day condition. We show an example in the supplementary video  where DRAN can not converge using (R+A), but successfully localizes with the full method (R+P+A), showing the increase in robustness. We can conclude that the final feature-metric optimization is able to refine the pose when the sparse-to-dense matching works well. The sparse-to-dense step allows convergence for wide baselines poses and filters out bad retrieval candidates.


Additionally, we experimented with a light version of DRAN (DRAN - light), and a hierarchical approach (GeM \cite{radenovic2016cnn} + S2DHM \cite{germain2019sparse} + PixLoc \cite{sarlin2021back}) which uses three different networks to perform retrieval, matching and pose refinement. This hierarchical approach gives similar results as our DRAN, while being slower (Table \ref{table:runtime}) and consuming more memory, having to load three networks. DRAN - light uses less layers for hypercolumns, limits points to 512 in matching and only refines the medium and fine features for the optimization, achieving a huge speed up with a very little drop in performance. That indicates that our approach could be further miniaturized \cite{sarlin2019coarse} and optimized to allow single network feature extraction in SLAM.
 
\begin{table}[t]
\resizebox{0.95\columnwidth}{!}{%
    \centering
    \begin{tabular}{lcccccc}
        \toprule
        Method & IR & Features & Match & Refine & \textit{N} & Total  \\
                                                                           
        \midrule
        DRAN (R+P+A)         & 0 & 265 & 1109 & 295 & 5 & 6.10\,s  \\
        DRAN - light        & 0 & 160 & 321 & 108 & 5 & 1.87\,s \\
        \midrule
        GeM+S2DHM+PixLoc & 66 & 302 & 1233 & 280 & 5 & 6.81\,s \\
        
        
        \bottomrule
    \end{tabular}}
    \vspace{-0.1cm}
    \caption{Runtime. Times shown in \,ms, except for Total.}
    \label{table:runtime}
    \vspace{-0.15cm}
\end{table}

Experiments were performed with an Intel® Core™ i7-10700K (3.80\,GHz) CPU and an Nvidia GeForce RTX 2080 Ti. We assume the database is already built, so we only extract features for the query image, match against \textit{N} reference images and refine the best candidate, so the total time is given by $t_{total} = t_{IR} + t_{features} + t_{match} \times N + t_{refine}$.


    


\section{Conclusions}


We have presented the first unified pipeline that performs all the tasks concerning Visual Localization under challenging conditions. DRAN can beat in performance the other Unified approaches in several datasets and conditions, using lower run time and memory budget than its counterpart hierarchical approach which employs different networks, while being competitive against complex Feature Matching pipelines. Convergence and robustness have increased due to the re-ranking step, while the feature-metric optimization is responsible for the final accuracy. We see DRAN as a first step towards the paradigm of a unique feature extractor, able to provide good features for feature-metric tracking and relocalization in SLAM. Despite its benefits, we are aware of the current performance gap between complex Feature Matching pipelines and Unified approaches. We believe that the joint training of retrieval and matching of unified features and the inclusion of more powerful matching methods in the initial pose estimate will close this gap.




\printbibliography

@inproceedings{germain2019sparse,
  title={Sparse-to-dense hypercolumn matching for long-term visual localization},
  author={Germain, Hugo and Bourmaud, Guillaume and Lepetit, Vincent},
  booktitle={2019 International Conference on 3D Vision (3DV)},
  pages={513--523},
  year={2019},
  organization={IEEE}
}

@inproceedings{sattler2018benchmarking,
  title={Benchmarking 6dof outdoor visual localization in changing conditions},
  author={Sattler, Torsten and Maddern, Will and Toft, Carl and Torii, Akihiko and Hammarstrand, Lars and Stenborg, Erik and Safari, Daniel and Okutomi, Masatoshi and Pollefeys, Marc and Sivic, Josef and others},
  booktitle={Proceedings of the IEEE conference on computer vision and pattern recognition},
  pages={8601--8610},
  year={2018}
}

@inproceedings{germain2020s2dnet,
  title={{S2Dnet}: learning image features for accurate sparse-to-dense matching},
  author={Germain, Hugo and Bourmaud, Guillaume and Lepetit, Vincent},
  booktitle={European Conference on Computer Vision},
  pages={626--643},
  year={2020},
  organization={Springer}
}

@article{paszke2019pytorch,
  title={Pytorch: An imperative style, high-performance deep learning library},
  author={Paszke, Adam and Gross, Sam and Massa, Francisco and Lerer, Adam and Bradbury, James and Chanan, Gregory and Killeen, Trevor and Lin, Zeming and Gimelshein, Natalia and Antiga, Luca and others},
  journal={Advances in neural information processing systems},
  volume={32},
  year={2019}
}

@inproceedings{deng2009imagenet,
  title={Imagenet: A large-scale hierarchical image database},
  author={Deng, Jia and Dong, Wei and Socher, Richard and Li, Li-Jia and Li, Kai and Fei-Fei, Li},
  booktitle={2009 IEEE conference on computer vision and pattern recognition},
  pages={248--255},
  year={2009},
  organization={Ieee}
}

@inproceedings{ronneberger2015u,
  title={U-net: Convolutional networks for biomedical image segmentation},
  author={Ronneberger, Olaf and Fischer, Philipp and Brox, Thomas},
  booktitle={International Conference on Medical image computing and computer-assisted intervention},
  pages={234--241},
  year={2015},
  organization={Springer}
}

@inproceedings{warburg2020mapillary,
  title={Mapillary street-level sequences: A dataset for lifelong place recognition},
  author={Warburg, Frederik and Hauberg, Soren and Lopez-Antequera, Manuel and Gargallo, Pau and Kuang, Yubin and Civera, Javier},
  booktitle={Proceedings of the IEEE/CVF conference on computer vision and pattern recognition},
  pages={2626--2635},
  year={2020}
}

@inproceedings{gladkova2021tight,
  title={Tight integration of feature-based relocalization in monocular direct visual odometry},
  author={Gladkova, Mariia and Wang, Rui and Zeller, Niclas and Cremers, Daniel},
  booktitle={2021 IEEE International Conference on Robotics and Automation (ICRA)},
  pages={9608--9614},
  year={2021},
  organization={IEEE}
}

@article{lepetit2009epnp,
  title={{EPnP}: An accurate o (n) solution to the {PnP} problem},
  author={Lepetit, Vincent and Moreno-Noguer, Francesc and Fua, Pascal},
  journal={International journal of computer vision},
  volume={81},
  number={2},
  pages={155},
  year={2009},
  publisher={Springer}
}

@article{GalvezTRO12,
  author={G{\'a}lvez-L{\'o}pez, Dorian and Tard{\'o}s, Juan D.},
  journal={IEEE Transactions on Robotics},
  title={Bags of Binary Words for Fast Place Recognition in Image Sequences},
  year={2012},
  month={October},
  volume={28},
  number={5},
  pages={1188--1197},
  doi={10.1109/TRO.2012.2197158},
  ISSN={1552-3098}
}

@INPROCEEDINGS{Nister2006,
  author={Nister, D. and Stewenius, H.},
  booktitle={2006 IEEE Computer Society Conference on Computer Vision and Pattern Recognition (CVPR'06)}, 
  title={Scalable Recognition with a Vocabulary Tree}, 
  year={2006},
  volume={2},
  number={},
  pages={2161-2168},
  doi={10.1109/CVPR.2006.264}}

@article{campos2020orb,
  title={{ORB-SLAM3}: An Accurate Open-Source Library for Visual, Visual--Inertial, and Multimap SLAM},
  author={Campos, Carlos and Elvira, Richard and Rodr{\'i}guez, Juan J G{\'o}mez and Montiel, Jos{\'e} MM and Tard{\'o}s, Juan D},
  journal={IEEE Transactions on Robotics},
  year={2021},
  publisher={IEEE}
}

@inproceedings{sarlin2021back,
  title={Back to the Feature: Learning Robust Camera Localization from Pixels to Pose},
  author={Sarlin, Paul-Edouard and Unagar, Ajaykumar and Larsson, Mans and Germain, Hugo and Toft, Carl and Larsson, Viktor and Pollefeys, Marc and Lepetit, Vincent and Hammarstrand, Lars and Kahl, Fredrik and others},
  booktitle={Proceedings of the IEEE/CVF Conference on Computer Vision and Pattern Recognition},
  pages={3247--3257},
  year={2021}
}

@article{yang2020ur2kid,
  title={{UR2KiD}: Unifying retrieval, keypoint detection, and keypoint description without local correspondence supervision},
  author={Yang, Tsun-Yi and Nguyen, Duy-Kien and Heijnen, Huub and Balntas, Vassileios},
  journal={arXiv preprint arXiv:2001.07252},
  year={2020}
}

@article{jia2014caffe,
  Author = {Jia, Yangqing and Shelhamer, Evan and Donahue, Jeff and Karayev, Sergey and Long, Jonathan and Girshick, Ross and Guadarrama, Sergio and Darrell, Trevor},
  Journal = {arXiv preprint arXiv:1408.5093},
  Title = {Caffe: Convolutional Architecture for Fast Feature Embedding},
  Year = {2014}
}

@article{von2020gn,
  title={{GN-Net}: The gauss-newton loss for multi-weather relocalization},
  author={Von Stumberg, Lukas and Wenzel, Patrick and Khan, Qadeer and Cremers, Daniel},
  journal={IEEE Robotics and Automation Letters},
  volume={5},
  number={2},
  pages={890--897},
  year={2020},
  publisher={IEEE}
}

@inproceedings{cao2020unifying,
  title={Unifying deep local and global features for image search},
  author={Cao, Bingyi and Araujo, Andre and Sim, Jack},
  booktitle={European Conference on Computer Vision},
  pages={726--743},
  year={2020},
  organization={Springer}
}

@inproceedings{sarlin2019coarse,
  title={From coarse to fine: Robust hierarchical localization at large scale},
  author={Sarlin, Paul-Edouard and Cadena, Cesar and Siegwart, Roland and Dymczyk, Marcin},
  booktitle={Proceedings of the IEEE/CVF Conference on Computer Vision and Pattern Recognition},
  pages={12716--12725},
  year={2019}
}

@article{xu2020deep,
  title={Deep probabilistic feature-metric tracking},
  author={Xu, Binbin and Davison, Andrew J and Leutenegger, Stefan},
  journal={IEEE Robotics and Automation Letters},
  volume={6},
  number={1},
  pages={223--230},
  year={2020},
  publisher={IEEE}
}

@inproceedings{lv2019taking,
  title={Taking a deeper look at the inverse compositional algorithm},
  author={Lv, Zhaoyang and Dellaert, Frank and Rehg, James M and Geiger, Andreas},
  booktitle={Proceedings of the IEEE/CVF Conference on Computer Vision and Pattern Recognition},
  pages={4581--4590},
  year={2019}
}

@inproceedings{noh2017large,
  title={Large-scale image retrieval with attentive deep local features},
  author={Noh, Hyeonwoo and Araujo, Andre and Sim, Jack and Weyand, Tobias and Han, Bohyung},
  booktitle={Proceedings of the IEEE international conference on computer vision},
  pages={3456--3465},
  year={2017}
}

@inproceedings{arandjelovic2016netvlad,
  title={{NetVLAD: CNN} architecture for weakly supervised place recognition},
  author={Arandjelovic, Relja and Gronat, Petr and Torii, Akihiko and Pajdla, Tomas and Sivic, Josef},
  booktitle={Proceedings of the IEEE conference on computer vision and pattern recognition},
  pages={5297--5307},
  year={2016}
}

@article{radenovic2018fine,
  title={Fine-tuning {CNN} image retrieval with no human annotation},
  author={Radenovi{\'c}, Filip and Tolias, Giorgos and Chum, Ond{\v{r}}ej},
  journal={IEEE transactions on pattern analysis and machine intelligence},
  volume={41},
  number={7},
  pages={1655--1668},
  year={2018},
  publisher={IEEE}
}

@inproceedings{rublee2011orb,
  title={{ORB}: An efficient alternative to {SIFT} or {SURF}},
  author={Rublee, Ethan and Rabaud, Vincent and Konolige, Kurt and Bradski, Gary},
  booktitle={2011 International conference on computer vision},
  pages={2564--2571},
  year={2011},
  organization={Ieee}
}

@article{lowe2004distinctive,
  title={Distinctive image features from scale-invariant keypoints},
  author={Lowe, David G},
  journal={International journal of computer vision},
  volume={60},
  number={2},
  pages={91--110},
  year={2004},
  publisher={Springer}
}

@inproceedings{dusmanu2019d2,
  title={D2-net: A trainable cnn for joint description and detection of local features},
  author={Dusmanu, Mihai and Rocco, Ignacio and Pajdla, Tomas and Pollefeys, Marc and Sivic, Josef and Torii, Akihiko and Sattler, Torsten},
  booktitle={Proceedings of the ieee/cvf conference on computer vision and pattern recognition},
  pages={8092--8101},
  year={2019}
}

@inproceedings{detone2018superpoint,
  title={Superpoint: Self-supervised interest point detection and description},
  author={DeTone, Daniel and Malisiewicz, Tomasz and Rabinovich, Andrew},
  booktitle={Proceedings of the IEEE conference on computer vision and pattern recognition workshops},
  pages={224--236},
  year={2018}
}

@inproceedings{sarlin2020superglue,
  title={Superglue: Learning feature matching with graph neural networks},
  author={Sarlin, Paul-Edouard and DeTone, Daniel and Malisiewicz, Tomasz and Rabinovich, Andrew},
  booktitle={Proceedings of the IEEE/CVF conference on computer vision and pattern recognition},
  pages={4938--4947},
  year={2020}
}

@article{mur2015orb,
  title={{ORB-SLAM}: a versatile and accurate monocular SLAM system},
  author={Mur-Artal, Raul and Montiel, Jose Maria Martinez and Tardos, Juan D},
  journal={IEEE transactions on robotics},
  volume={31},
  number={5},
  pages={1147--1163},
  year={2015},
  publisher={IEEE}
}

@article{engel2017direct,
  title={Direct sparse odometry},
  author={Engel, Jakob and Koltun, Vladlen and Cremers, Daniel},
  journal={IEEE transactions on pattern analysis and machine intelligence},
  volume={40},
  number={3},
  pages={611--625},
  year={2017},
  publisher={IEEE}
}

@inproceedings{von2020lm,
  title={{LM-Reloc}: Levenberg-Marquardt based direct visual relocalization},
  author={Von Stumberg, Lukas and Wenzel, Patrick and Yang, Nan and Cremers, Daniel},
  booktitle={2020 International Conference on 3D Vision (3DV)},
  pages={968--977},
  year={2020},
  organization={IEEE}
}

@inproceedings{teichmann2019detect,
  title={Detect-to-retrieve: Efficient regional aggregation for image search},
  author={Teichmann, Marvin and Araujo, Andre and Zhu, Menglong and Sim, Jack},
  booktitle={Proceedings of the IEEE/CVF Conference on Computer Vision and Pattern Recognition},
  pages={5109--5118},
  year={2019}
}

@inproceedings{revaud2019learning,
  title={Learning with average precision: Training image retrieval with a listwise loss},
  author={Revaud, Jerome and Almaz{\'a}n, Jon and Rezende, Rafael S and Souza, Cesar Roberto de},
  booktitle={Proceedings of the IEEE/CVF International Conference on Computer Vision},
  pages={5107--5116},
  year={2019}
}

@inproceedings{sun2021loftr,
  title={{LoFTR}: Detector-free local feature matching with transformers},
  author={Sun, Jiaming and Shen, Zehong and Wang, Yuang and Bao, Hujun and Zhou, Xiaowei},
  booktitle={Proceedings of the IEEE/CVF Conference on Computer Vision and Pattern Recognition},
  pages={8922--8931},
  year={2021}
}

@inproceedings{schonberger2016structure,
  title={Structure-from-motion revisited},
  author={Schonberger, Johannes L and Frahm, Jan-Michael},
  booktitle={Proceedings of the IEEE conference on computer vision and pattern recognition},
  pages={4104--4113},
  year={2016}
}

@inproceedings{li2018megadepth,
  title={Megadepth: Learning single-view depth prediction from internet photos},
  author={Li, Zhengqi and Snavely, Noah},
  booktitle={Proceedings of the IEEE Conference on Computer Vision and Pattern Recognition},
  pages={2041--2050},
  year={2018}
}

@inproceedings{engel2014lsd,
  title={{LSD-SLAM}: Large-scale direct monocular SLAM},
  author={Engel, Jakob and Sch{\"o}ps, Thomas and Cremers, Daniel},
  booktitle={European conference on computer vision},
  pages={834--849},
  year={2014},
  organization={Springer}
}

@inproceedings{newcombe2011dtam,
  title={{DTAM}: Dense tracking and mapping in real-time},
  author={Newcombe, Richard A and Lovegrove, Steven J and Davison, Andrew J},
  booktitle={2011 international conference on computer vision},
  pages={2320--2327},
  year={2011},
  organization={IEEE}
}

@article{davison2007monoslam,
  title={{MonoSLAM}: Real-time single camera SLAM},
  author={Davison, Andrew J and Reid, Ian D and Molton, Nicholas D and Stasse, Olivier},
  journal={IEEE transactions on pattern analysis and machine intelligence},
  volume={29},
  number={6},
  pages={1052--1067},
  year={2007},
  publisher={IEEE}
}

@inproceedings{klein2007parallel,
  title={Parallel tracking and mapping for small AR workspaces},
  author={Klein, Georg and Murray, David},
  booktitle={2007 6th IEEE and ACM international symposium on mixed and augmented reality},
  pages={225--234},
  year={2007},
  organization={IEEE}
}

@inproceedings{radenovic2016cnn,
  title={{CNN} image retrieval learns from {BoW}: Unsupervised fine-tuning with hard examples},
  author={Radenovi{\'c}, Filip and Tolias, Giorgos and Chum, Ond{\v{r}}ej},
  booktitle={European conference on computer vision},
  pages={3--20},
  year={2016},
  organization={Springer}
}

@inproceedings{kendall2017geometric,
  title={Geometric loss functions for camera pose regression with deep learning},
  author={Kendall, Alex and Cipolla, Roberto},
  booktitle={Proceedings of the IEEE conference on computer vision and pattern recognition},
  pages={5974--5983},
  year={2017}
}

@inproceedings{naseer2017deep,
  title={Deep regression for monocular camera-based 6-dof global localization in outdoor environments},
  author={Naseer, Tayyab and Burgard, Wolfram},
  booktitle={2017 IEEE/RSJ International Conference on Intelligent Robots and Systems (IROS)},
  pages={1525--1530},
  year={2017},
  organization={IEEE}
}

@article{leyvavallina2021gcl,
  title={Generalized Contrastive Optimization of Siamese Networks for Place Recognition}, 
  author={María Leyva-Vallina and Nicola Strisciuglio and Nicolai Petkov},
  journal={arXiv preprint arXiv:2103.06638},
  year={2021},
  url={https://arxiv.org/abs/2103.06638}
}

\end{document}